\documentclass[11pt]{article}
\usepackage{coling2020}
\usepackage{times}
\usepackage{url}
\usepackage{latexsym}
\usepackage{hyperref}
\usepackage{graphicx}
\usepackage{amsmath}
\usepackage{amssymb}
\usepackage{float}

\title{Learning Efficient Task-Specific Meta-Embeddings with Word Prisms}

\author{Jingyi He\thanks{\; Equal contribution. $\dagger$ This work was pursued prior to Kian's employment at BMO. }\textsuperscript{\; 1} \and 
KC Tsiolis\footnotemark[1]\textsuperscript{\; 1} \and 
Kian Kenyon-Dean$^{\dagger}$\footnotemark[1]\textsuperscript{\; 2} \and Jackie C. K. Cheung\textsuperscript{1}\\
Mila -- Qu{\'e}bec AI Institute / McGill University, Montr{\'e}al, QC, Canada \textsuperscript{1}\\
BMO AI Capabilities Team -- Bank of Montr{\'e}al, Toronto, ON, Canada \textsuperscript{2}\\
 {\tt \{jingyi.he,kc.tsiolis\}@mail.mcgill.ca} \and \tt {jcheung@cs.mcgill.ca}\\
 \tt {kian.kenyon-dean@bmo.com}
 }

\date{}
\date{}

\begin{document}
\maketitle
\begin{abstract}
Word embeddings are trained to predict word cooccurrence statistics, which leads them to possess different lexical properties (syntactic, semantic, etc.) depending on the notion of context defined at training time.
These properties manifest when querying the embedding space for the most similar vectors, and when used at the input layer of deep neural networks trained to solve downstream NLP problems.
Meta-embeddings combine multiple sets of differently trained word embeddings, and have been shown to successfully improve intrinsic and extrinsic performance over equivalent models which use just one set of source embeddings. 
We introduce \textit{word prisms}: a simple and efficient meta-embedding method that learns to combine source embeddings according to the task at hand. 
Word prisms learn orthogonal transformations to linearly combine the input source embeddings, which allows them to be very efficient at inference time.
We evaluate word prisms in comparison to other meta-embedding methods on six extrinsic evaluations and observe that word prisms offer improvements in performance on all tasks.\footnote{\url{https://github.com/kylie-box/word\_prisms}}


\end{abstract}

\section{Introduction}
\label{1_introduction}



\blfootnote{This work is licensed under a Creative Commons Attribution 4.0 International Licence. Licence details: http://creativecommons.org/licenses/by/4.0/.}

A popular approach to representing word meaning in NLP is to characterize a word by ``the company that it keeps'' \cite{firth1957synopsis}. This intuition is the basis of famous word embedding techniques such as Word2vec \cite{mikolov2013efficient} and Glove \cite{pennington2014glove}.
However, the question of \textit{what company a word keeps} --- i.e., what should define a word's context --- is open. A word's context could be defined via a symmetric window of 1, 2, 5, 10, 20 words, the words that precede it, the words that follow it, the words with which it shares a dependency edge, etc. Determining the utility of such different notions of context for training word embeddings is a problem that has attracted considerable attention \cite{yatbaz2012learning,levy2014dependency,bansal2014,lin2015unsupervised,melamud2016role,lison2017redefining} but there is no conclusive evidence that any single notion of context could be the best for solving NLP problems in general. 
Thus, many deep learning solutions for NLP have yet another hyperparameter to tune: what set of word embeddings should be selected for the input layer of the model. As NLP tasks become more and more complex, the practice of providing a deep model with only one notion of a word's meaning becomes limiting.


Word meta-embeddings address aspects of this problem by proposing techniques for combining multiple sets of word embeddings before providing them into the input layer of a downstream model.
%
\newcite{yin2016learning} motivated word meta-embeddings by arguing that they are advantageous for the following reasons: \textit{diversity} --- combining embeddings trained with different algorithms on different corpora will allow for more distinct meanings of the words to persist; and, \textit{coverage} --- combining embeddings trained on different corpora help to better solve the out-of-vocabulary problem. 
However, they did not acknowledge that different sets of word embeddings can be diverse even when trained on the same corpus with the same algorithm, so long as their context windows are different.
Additionally, due to the various practical and theoretical similarities between different algorithms \cite{levy2014neural,levy2015improving,newell2019deconstructing}, the gains to be found in diversifying at the level of algorithmic variation are likely to be minimal.
With regard to vocabulary coverage, the out-of-vocabulary problem is at least partially addressed by character n-gram based embedding algorithms such as FastText \cite{joulin2017bag} and subword-based decomposition techniques that can be applied post-training \cite{zhao2018generalizing,sasaki2019subword}.  
Nonetheless, meta-embeddings have been shown to consistently outperform models that use only a single set of embeddings in their input layer. Our goal is to determine how to best combine many sets of input embeddings in order to obtain high quality results in downstream tasks.

\begin{figure}
    \centering
    \includegraphics[trim=0 150 0 0 ,width=\linewidth]{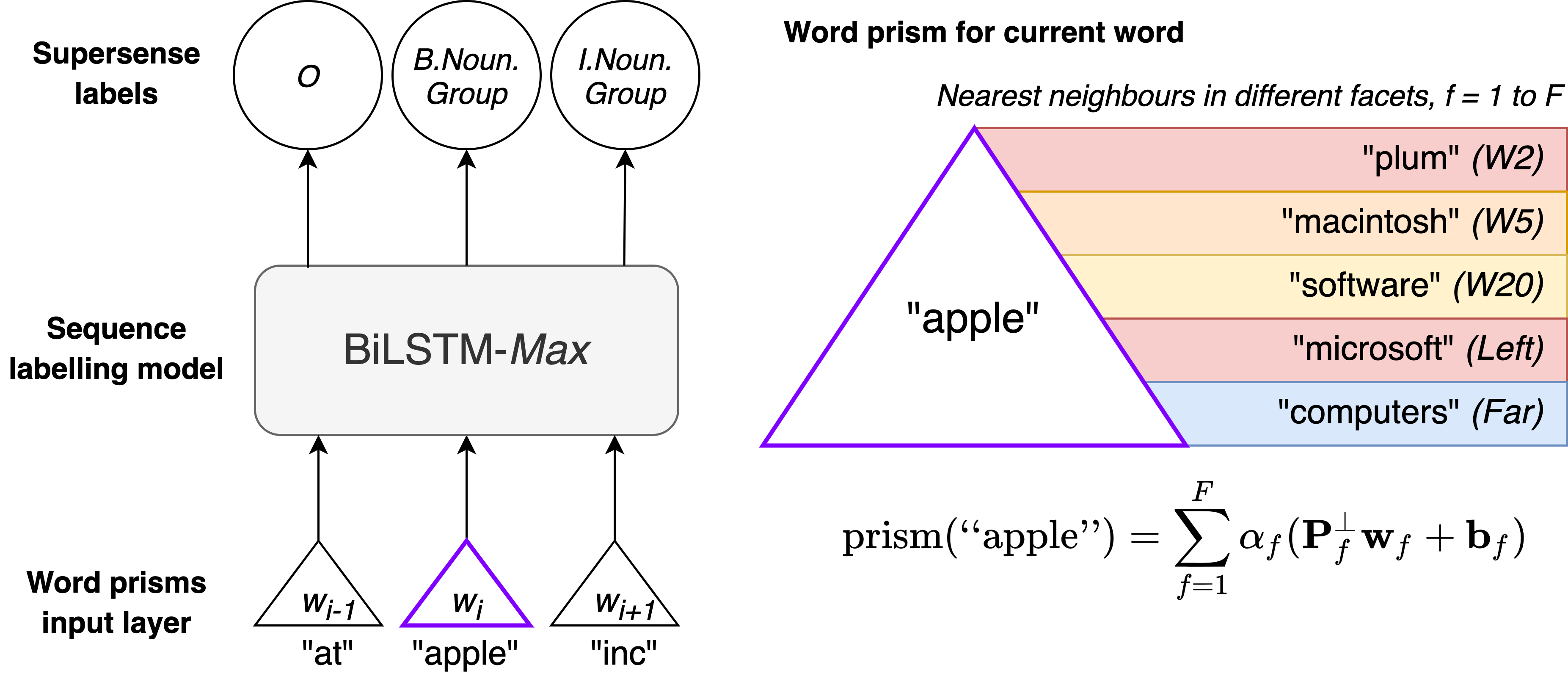}
    \caption{Word prisms (§\ref{3_method}) during supersense tagging (§\ref{4_experiments}). We display the nearest neighbor to the embedding for the word ``apple'' in five of the window-based facets (Figure~\ref{fig:windows}, §\ref{sec:facet-combos}) used in the downstream model (§\ref{4_experiments}). We include the $\perp$ symbol on the learned transformation, $\textbf{P}_f$, to indicate its orthogonality.}
    \label{fig:wordprism}
\end{figure}

This work proposes \textit{word prisms} as a simple and general way to produce and understand meta-embeddings, visualized in Figure~\ref{fig:wordprism}.
Word prisms excel at combining many sets of source embeddings, which we call \textit{facets}. They do so by learning task-specific orthogonal transformations to map embeddings from their facets to the common meta-embedding space. This produces a vector space that is more disentangled than the original space of facet embeddings. It allows the combination of multiple source embeddings while preserving most information within each embedding set.

 To our knowledge, this work is the first to incorporate both explicit orthogonal transformations of source embeddings and importance weights for source embedding sets that are dynamically learned with the downstream tasks in the same meta-embedding method. Furthermore, it is the first to explore combining so many sets of source embeddings (thirteen).
We compare the word prisms method to other standard meta-embedding algorithms (averaging \cite{coates2018frustratingly}, concatenation \cite{yin2016learning}, and dynamic meta-embeddings, DMEs, \cite{kiela2018dynamic}). 
Word prisms overcome the shortcomings of each of these algorithms:
(1) in averaging, performance deteriorates considerably when there are many facets --- the orthogonal transformations in word prisms resolve this problem; 
(2) concatenation and DMEs are too expensive during inference when there are many facets --- word prisms only need the final meta-embeddings at inference time, making them as efficient as averaging.
Our results demonstrate that neural downstream models using word prisms generally obtain better results than the other algorithms across six downstream tasks, including supersense tagging, POS tagging, named entity recognition, natural language inference, and sentiment analysis. 
In our ablation studies, we find that our method improves performance on downstream tasks even when the vocabulary is the same across nine source facets trained on the same corpus that differ only by the definition of context window (see Figure~\ref{fig:windows}); performance further improves by incorporating four more sets of off-the-shelf facets.

\section{Related Work}
\label{2_related_work}

Pre-trained word embedding algorithms use word cooccurrence statistics from a training corpus to map words to a low-dimensional vector space such that words with similar meanings are mapped to similar points in vector space. However, changing the embedding algorithm, training corpus, or definition of cooccurrence can have a strong impact on the resulting embeddings. Consequently, word \textit{meta-embedding} algorithms have been developed to combine multiple embedding sets.

\subsection{Word embedding training and notions of context}

Word embeddings are trained to reflect the cooccurrence statistics in the input corpus, which depend on the specific definition of context being employed.
The standard definition in Word2vec \cite{mikolov2013efficient} and GloVe \cite{pennington2014glove} is a symmetric context window of fixed size around each word in the corpus. \newcite{levy2014dependency} explore dependency-based contexts, where the context of each word is defined as its governor and dependents, along with the corresponding dependency relation labels. 
They observe that embeddings trained on smaller context windows and dependency-based contexts relate words that can be substituted for one another. By contrast, embeddings trained on larger context windows relate words which address the same topic.
\newcite{bansal2014} observe the same phenomenon.
\newcite{lin2015unsupervised} find that small window sizes work best for POS tagging. \newcite{lison2017redefining} evaluate word similarity and word analogy task performance for SGNS embeddings trained on different window sizes, as well as left-sided and right-sided context windows. 


\begin{figure}
    \centering
    \includegraphics[trim=0 20 0 0,width=\linewidth]{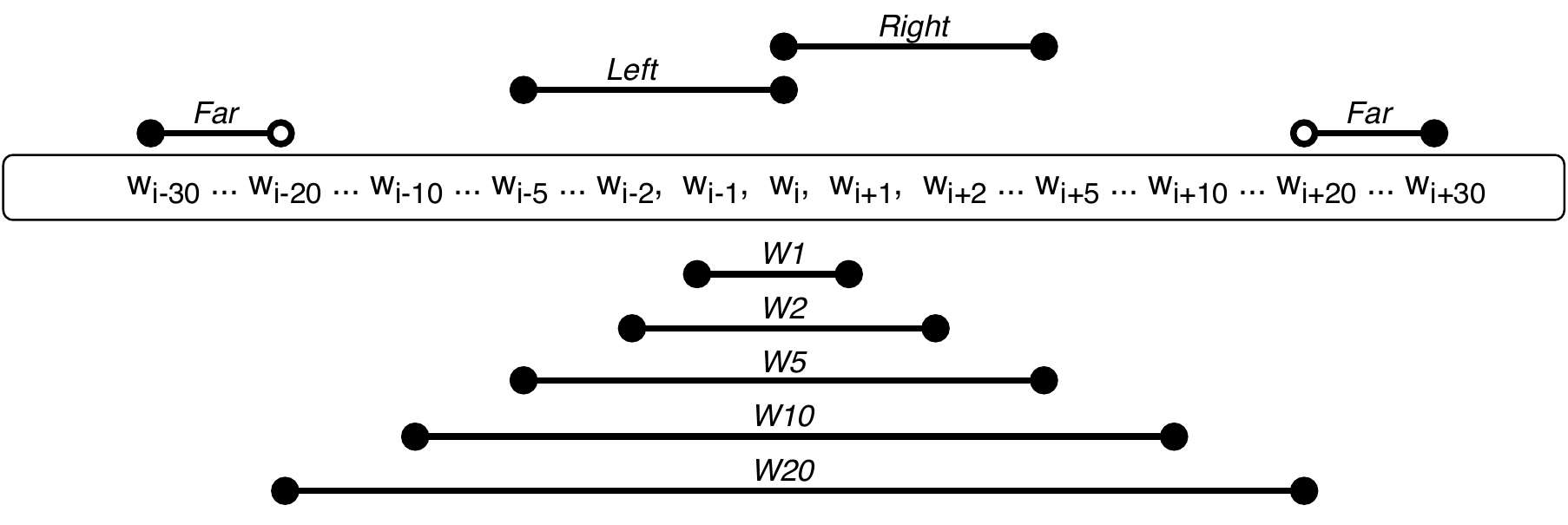}
    \caption{The eight different context windows with which we trained word embedding facets (note that the dependency-based facet is not included here). Word $w_i$ represents the center of the context windows.}
    \label{fig:windows}
\end{figure}

\subsection{Word meta-embeddings}
Previous work has shown that combining embedding sets can lead to improvements in downstream performance. \newcite{melamud2016role} combine embeddings trained with different notions of context via concatenation, as well as via SVD and CCA, leading to improved performance in multiple downstream tasks. However, they only combine two embedding sets at a time. \newcite{yin2016learning} introduce the term ``meta-embeddings" and demonstrate that concatenation and singular value decomposition (SVD) are solid baselines on word similarity, analogy, and POS tagging tasks. They propose \textsc{1toN}, which simultaneously learns meta-embeddings and projections from the meta-embedding space to each individual source embedding space. \newcite{ghannay2016word} apply PCA and autoencoders after concatenating source embeddings. \newcite{zhang2016mgnc} apply a convolutional layer to each source embedding before concatenating the resulting feature maps. \newcite{bollegala2018think} represent the meta-embedding for a word as a linear combination of the meta-embeddings for its nearest neighbours in each source embedding set. \newcite{bollegala2018learning} produce meta-embeddings by either averaging or concatenating the outputs of encoders which take GloVe and CBOW \cite{mikolov2013efficient} embeddings as input. 

\newcite{coates2018frustratingly} demonstrate that, in certain settings, meta-embeddings produced by averaging can be as performant as concatenated ones.
\newcite{kiela2018dynamic} propose dynamic meta-embeddings (DMEs), which perform attention over the linearly transformed source embeddings. The linear transformations applied to the source embeddings are not constrained to be orthogonal. Their model learns which source embedding sets are most useful for a particular downstream task and for particular classes of words. They also present a contextualized version, where attention weights also depend on a word's surrounding context, but this provides little to no improvement on their downstream evaluations. By contrast, we propose a simpler attention mechanism by learning a single importance weight for each source embedding set, and we apply orthogonal transformations to source embeddings prior to linear combination. We also experiment with a larger selection of source embeddings, including embeddings trained with different notions of context.

Orthogonal transformations have previously been employed in the context of mapping monolingual embeddings for different languages into a common space \cite{artetxe-etal-2016-learning,smith2017offline,artetxe2018generalizing,conneau2018word,doval-etal-2018-improving}. With these alignment transformations on mono-lingual space, one can obtain a better cross-lingual integration of the vector spaces. Recent work has also found that applying orthogonal transformations to source embeddings facilitates averaging \cite{garcia2020common,jawanpuria-etal-2020-learning}. We expand on this work by incorporating orthogonal transformations in word prisms, which learn word meta-embeddings for specific downstream tasks. Additionally, we provide an analysis of source embeddings before and after orthogonal transformation, which leads to the insight that these mappings cause source embedding sets to be more easily clusterable within the meta-embedding space.


\section{Word Prisms and Meta-Embeddings}
\label{3_method}

In this section we introduce meta-embeddings, word prisms, and the source embeddings they are composed with in this work. A meta-embedding combines pre-trained embeddings from multiple sources (e.g., from Glove and FastText), which we call \textit{facets}. We define  $\mathbf{w}_f$ as the embedding of word $w$ in facet $f$, for each facet $f \in \{1, ..., F\}$. The dimensionality of each embedding in a facet is denoted $d_f$, and the final dimensionality of the meta-embedding is $d'$. The following equation represents the general form of all meta-embedding variants and baselines considered in this work:
\begin{align} \label{eq:meta}
    \operatorname{meta}(w) &= \sum_{f=1}^{F} \alpha_{f} (\textbf{P}_f \textbf{w}_{f} + \textbf{b}_f)  \qquad
    \text{s.t.} \,\, \alpha_{f} \in \mathbb{R}, \, \textbf{w}_{f} \in \mathbb{R}^{d_f}, \, \textbf{P}_f \in \mathbb{R}^{d' \times d_f}, \, \textbf{b}_f \in \mathbb{R}^{d'}.
\end{align}

That is, the meta-embedding for a word $w$, $\operatorname{meta}(w) \in \mathbb{R}^{d'}$ is constructed as follows: first, it is projected by a linear transformation (learned or fixed) characterized by a matrix $\textbf{P}_f$ and bias $\textbf{b}_f$ for some set of embedding facets. Next, the meta-embedding is a linear combination of the transformed embeddings scaled by some (learned or fixed) weights, $\alpha_f$.
%
The vocabulary of a word prism is the union of the vocabularies of its facets. If a word is out-of-vocabulary for a facet $f$, we assign its representation $\textbf{w}_f$ to be the centroid of the embeddings in the facet.

\subsection{Word prisms}

Word prisms learn orthogonal transformation matrices $\textbf{P}_f$ and bias vectors $\textbf{b}_f$ via back-propagation to make the space of input facets more well-separated so that the downstream model can learn which lexical qualities in the facets are most appropriate for the given task at hand.
It is desirable to impose an orthogonality constraint on the transformation matrix because orthogonal matrices preserve the dot products within the original vector space, which has been shown to be important in studies of multilingual embeddings \cite{artetxe-etal-2016-learning}. 
This requires $\textbf{P}_f$ to be square ($\textbf{P}_f \in \mathbb{R}^{d' \times d'}$) which further requires the dimensionality of the facets to all be the same (i.e., each $d_f = d'$, without loss of generality\footnote{When dimensions are different, simple strategies can be pursued to equalize them. For example, zero-padding short embeddings \cite{coates2018frustratingly} or using SVD to compress long embeddings are reliable strategies.}).
After each gradient descent update, we apply the following update rule used by \newcite{cisse2017parseval} and \newcite{conneau2018word}, which approximates a procedure that keeps each $\mathbf{P_f}$ on the manifold of orthogonal matrices:
\begin{align} \label{eq:orthog}
    \mathbf{P_f} \leftarrow(1+\beta) \mathbf{P_f}-\beta\left(\mathbf{P_f} \mathbf{P_f}^{T}\right) \mathbf{P_f}.
\end{align}
The orthogonal transformation keeps the L2 norm of the original embeddings the same, since it does not rescale the vectors. Our preliminary experiments found $\beta = 0.001$ is a good option.

Word prisms also learn a linear combination of the projected facets to further adapt to the task at hand. Indeed, it is necessary to learn the $\alpha_f$ separately from $\textbf{P}_f$ since the transformations are orthogonal and cannot perform rescaling independently. 
Word prisms learn the facet-level weight coefficients, $\alpha_f$, directly, where each $\alpha_f$ is a floating-point number also learned via back-propagation from the downstream task signal, initialized to be $1/F$ for each facet. That is, the parameters in word prisms are learned simultaneously with the downstream model for a given task. This approach is advantageous because it allows the model to assign importance weights to each facet, but it is not bound to do so via a dynamic attention vector. So, for word prisms, all of the meta-embeddings can be pre-computed for a vocabulary after training. This means that a word prism model in an inference-only production environment benefits from low memory complexity, as it does not need to hold all of the original facets in memory, only the meta-embeddings. Thus, given a vocabulary size $V$, and number of facets $F$, the memory complexity during inference is only $O(V)$ for word prisms (and the average baseline), but is $O(VF)$ for DMEs (and the concatenation baseline).

\subsection{Facets} \label{sec:facet-combos}
We include 13 various facets into word prisms with the aim of capturing a wide variety of semantic and syntactic information. To our knowledge, this is the first work on meta-embeddings to explore combining so many sets of source embeddings.
Our collection of facets is diverse in the following two ways: (1) it incorporates many notions of context using the same algorithm and the same corpus; (2) it incorporates off-the-shelf embeddings trained on much larger corpora and tuned to knowledge graphs. 

We make use of nine different notions of context to train standard PMI-based word embeddings \cite{levy2014neural,newell2019deconstructing}, each with a dimension of 300 and vocabulary size of 500,000. Training is done with the open-source sampling-based implementation of Hilbert-MLE\footnote{\url{https://github.com/enewe101/hilbert}} \cite{newell2019deconstructing}, which facilitates the use of arbitrarily structured context windows for training.
For eight of the nine notions of context, embeddings are trained on the Gigaword 3 corpus ~\cite{gigawords3} combined with a Wikipedia 2018 dump, which amounts to approximately 6 billion tokens. 

We visualize the different window settings for these eight embedding sets in Figure~\ref{fig:windows}. Letting \textit{W} be the window size, we trained the following sets of embeddings: \textit{W1}, \textit{W2}, \textit{W5}, \textit{W10}, and \textit{W20}. Furthermore, we trained embeddings using only a \textit{Left} context of 5 words, and another set of embeddings with only a \textit{Right} context of 5 words. Lastly, we trained a set of embeddings with only a \textit{Far} context window, which only includes words between 20 and 30 words away, in order to create strong topic-based representations. We also trained a variant of dependency-based embeddings (\textit{Deps}) \cite{levy2014dependency}, where we defined a word's context to be its governor. We ran the CoreNLP~\cite{manning-etal-2014-stanford} dependency parser on Gigaword 3 to obtain a parsed corpus. 
%
%

We also experiment with the following off-the-shelf embeddings: \textbf{GloVe} \cite{pennington2014glove}; trained on 840B tokens from the Common Crawl Corpus with 2.2M words in the vocabulary. \textbf{FastText} \cite{joulin2017bag}; trained on 600B tokens from the Common Crawl Corpus with 2M words in the vocabulary. \textbf{ConceptNet Numberbatch} \cite{speer2017conceptnet}; retrofitted \cite{faruqui2015retrofitting} on both Word2vec~\cite{mikolov2013distributed} and GloVe~\cite{pennington2014glove} with 516K words in the vocabulary; this facet allows us to incorporate information from knowledge graphs. \textbf{LexSub} \cite{lexsub}; GloVe embeddings trained on 6B tokens from Wikipedia 2014 and the Gigaword 5 corpus \cite{gigaword5}, modified so that they can easily be projected into ``lexical subspaces", in which a word's nearest neighbours reflect a particular lexical relation (e.g. synonymy, antonymy, hypernymy, meronymy).

\section{Experiments}
\label{4_experiments}
Our experiments seek to determine if: (1) word prisms offer improvements over the other common meta-embedding methods; and, (2) if it is desirable to produce meta-embeddings with many different notions of context from the same corpus.  
For (1), we pursue a variety of experiments comparing word prisms to the following meta-embedding methods: the averaging baseline, the concatenation baseline, and dynamic meta-embeddings (DMEs) \cite{kiela2018dynamic}. For (2), we experiment with several sets of meta-embedding facet combinations. The first set is FastText and Glove (\textbf{FG}), as is done by \newcite{kiela2018dynamic}. The second set is a combination of 13 different facets (\textbf{All}), as detailed in §\ref{sec:facet-combos}. In §\ref{5_results} we present our main results, and in §\ref{6_discussion} we present an ablation study to determine the impact of other meta-embedding combinations and the transformation matrices in word prisms.

\subsection{Baselines}
We will compare word prisms with three baseline algorithms. The first two, averaging and concatenation, are standard meta-embedding methods often explored in studies on meta-embeddings \cite{yin2016learning,coates2018frustratingly}. The third is dynamic meta-embeddings (DMEs) \cite{kiela2018dynamic}.

\paragraph{Average baseline.} Averaging word embeddings is the simplest method to create meta-embeddings.  Assuming each facet dimension $d_f$ is equal to $d'$ (as with word prisms), this baseline corresponds to Equation~\ref{eq:meta} with the following fixed parameter settings: $\alpha_f = \frac{1}{F}$, $\textbf{P}_f = \textbf{I}_{d'}$, and $\textbf{b}_f = \vec{0}$.

Averaging is a sensible strategy to combine multiple source word embeddings, since, first of all, it aggregates the information from all the input facets without introducing additional parameters. Second, it captures semantic information by preserving the relative word distances within the embedding spaces~\cite{coates2018frustratingly}. However, as we demonstrate later (§\ref{5_results}), the quality of this baseline deteriorates when there are many different facets as the signals start to become too mixed.

\paragraph{Concatenation baseline.} Concatenating multiple source embeddings is another trivial way to construct meta-embeddings. The parameter settings here correspond to Equation~\ref{eq:meta} when $d' = \sum_{f=1}^{F} d_f$, $\alpha_f = 1$, $\textbf{b}_f = \vec{0}$, and $\textbf{P}_f$ is a fixed selector matrix 
that places embeddings into their corresponding concatenated positions; more simply, $\operatorname{meta}(w) = [ \textbf{w}_1, \ldots, \textbf{w}_F ]$.

Concatenation can be desirable because it maintains all of the structure of the original embeddings. However, it is problematic because the dimensionality increases linearly with respect to the number of facets, requiring more model parameters to be learned at the input layer for downstream model.


\paragraph{Dynamic meta-embeddings.}
\newcite{kiela2018dynamic} introduced DMEs, demonstrating that sentence representations can be improved by combining multiple source embeddings with dynamically learned linear transformations and attention weights. 
Note that, like with concatenation, it is necessary to maintain all of the individual facets in memory during inference when using DMEs.  

DMEs are encapsulated in Equation~\ref{eq:meta} via the following parameter settings: $d'$ is a hyperparameter for the desired meta-embedding size (set to $256$ by \newcite{kiela2018dynamic});
$\textbf{P}_f$ and $\textbf{b}_f$ are learned via backpropagation from supervised learning during the current task; the $\alpha_f$ are obtained via a self-attention mechanism on an additional learned parameter vector $\textbf{a} \in \mathbb{R}^{d'}$: $\alpha_f = \phi(\textbf{a} \cdot (\textbf{P}_f \textbf{w}_f + \textbf{b}_f) + \textit{b})$, where $\phi$ is the softmax function and $\textit{b} \in \mathbb{R}$ is an additional learned bias parameter.

\subsection{Datasets and downstream models}
We evaluate meta-embedding methods on a variety of downstream text classification and sequence labelling tasks. For text classification, we choose the Stanford Sentiment Treebank binary sentiment analysis dataset (\textbf{SST2})~\cite{socher-etal-2013-recursive} and the Stanford NLI~\cite{snli:emnlp2015} (\textbf{SNLI}) benchmark. For sequence labelling, we select the CoNLL 2003 named entity recognition task (\textbf{NER})~\cite{Tjong_Kim_Sang_2003}, POS tagging on the Brown corpus (\textbf{Brown})\footnote{Retrieved from the NLTK toolkit: \url{http://www.nltk.org/nltk\_data/}.}, POS tagging on the WSJ corpus (\textbf{WSJ})~\cite{marcus1993building}, and Supersense tagging~\cite{ciaramita2003supersense} on the Semcor 3.0 corpus (\textbf{Semcor})~\cite{miller1993semantic}. 
Supersense tagging is a problem situated between NER and word sense disambiguation. The task consists of 41 lexicographer class labels for nouns and verbs with IOB tags, producing 83 fine-grained classes in total. We report the micro F1 score for the supersense tagging and the NER tagging tasks, discarding the O-tags in the predictions, as is standard \cite{alonso2017multitask,changpinyo2018multi}. For the rest of the tasks, we report the accuracy on the test set. We use the standard train-validation-test splits whenever they are provided with the dataset. Otherwise, we split 10\% of the training set to be the validation set for hyperparameter tuning.

We use a simple neural model to compare meta-embedding methods. To replicate the models used by \newcite{kiela2018dynamic}, we use a single layer BiLSTM with $512$ hidden units in SST2 and $1024$ hidden units in SNLI for sentence encoding. The sentence representations are learned by max pooling over the forward and backward hidden states. We use the following representation for a pair of hypothesis and premise: $[\textbf{u}, \textbf{v}, \textbf{u} * \textbf{v}, |\textbf{u - v}|]$. The $*$ operator denotes the element-wise multiplication. The sentence encoder is followed by a $512$ dimensional MLP with ReLU activation.
For all the sequence labelling tasks, we use a 2-layer BiLSTM with $256$ hidden units for sequence encoding. 

We use standard cross entropy loss for all supervised downstream tasks. The parameters of the downstream model and word prisms are learned via back-propagation. We choose the initial learning rate to be $0.001$ for sequence labeling tasks and $0.0004$ for text classification tasks with a reduction factor of $0.1$ if there is no improvement on the validation set after 2 consecutive epochs. In all of our experiments, we keep the source embeddings in their original forms without performing normalization. Counterintuitively, normalizing the source embeddings to have unit norm makes little difference in the text classification tasks but substantially hurts the performance of the sequence labelling tasks. Previous work~\cite{schakel2015measuring} shows the length of the embedding vector encodes the unigram frequency of the word, which is useful in the sequence labelling tasks.



\section{Results}
\label{5_results}

\begin{table*}[t]
    \centering
        \begin{tabular}{c c c c c c c c c c c c c}
            \hline
            Model & Facets & Semcor & WSJ & Brown & NER & SNLI & SST2\\
            \hline
            Average & FG & 69.42 $\pm$ .1 & 96.76 $\pm$ .02 & 98.44 $\pm$ .02 & 90.16 $\pm$ .2 & 85.33 $\pm$ .3 & 87.76  $\pm$ .5 \\
            Concat & FG & 72.23 $\pm$ .2 & 96.85 $\pm$ .04 & 98.53 $\pm$ .02 & 90.49 $\pm$ .1 & 85.45 $\pm$  .3 & \textbf{88.57} $\pm$ .3\\
            DME & FG & 72.15 $\pm$ .2 & 96.81 $\pm$ .03 & 98.53 $\pm$ .02 & 89.49 $\pm$ .2 & 85.57 $\pm$ .3 & 88.10 $\pm$ .6\\
            Prism & FG & 73.51 $\pm$ .1 & 96.91 $\pm$ .01 & 98.58 $\pm$ .02 & \underline{90.70} $\pm$ .4 & \textbf{85.82} $\pm$ .1 & 87.80 $\pm$ .6\\
            \hline
            Average & All & 65.34 $\pm$ .3 & 96.63 $\pm$ .01 & 98.21 $\pm$ .03 & 88.92 $\pm$ .3 & 83.91 $\pm$ .1 & 86.03 $\pm$ .9 \\
            Concat & All & \textbf{73.95} $\pm$ .1 & \underline{97.02} $\pm$ .01 & \underline{98.63} $\pm$ .01 & 90.55 $\pm$ .1 & 84.03 $\pm$ .2 & 88.15 $\pm$ .2\\
            DME & All & 72.09 $\pm$ .1 & 96.89 $\pm$ .01 & 98.58 $\pm$ .02 & 89.36 $\pm$ .3 & 85.47 $\pm$ .1 & 87.63 $\pm$ .6\\
            Prism & All & \underline{73.82} $\pm$ .2 & \textbf{97.04} $\pm$ .01 & \textbf{98.65} $\pm$ .01 & \textbf{90.74} $\pm$ .2 & \underline{85.71} $\pm$ .3 & \underline{88.45} $\pm$ .5\\
            \hline
        \end{tabular}
    \caption{Test set results for word prisms and baseline meta-embedding algorithms (concatenation, averaging, and DMEs) on different combinations of input facets (§\ref{sec:facet-combos}) --- \textbf{FG} is FastText and Glove only, \textbf{All} is all 13 facets. We report the mean and standard deviation from runs with five different random seeds. Best result is \textbf{bold}, second best is \underline{underlined}.} 
    \label{tab:prism_vs_others}
\end{table*}

Table~\ref{tab:prism_vs_others} presents the main results for this work on four sequence labelling tasks and two text classification tasks. We compare word prisms to standard meta-embedding baselines (averaging, concatenation) and dynamic meta-embeddings (DME). In the first four rows, we experiment with FastText and Glove (\textbf{FG}) as a two-facet combination, while the next four rows use a 13-facet combination (\textbf{All}) detailed in §\ref{sec:facet-combos}.

Our first finding is that word prisms almost always offer substantial improvements over DMEs, regardless of whether we are using two facets or thirteen. The only exception is in the case of text classification on SST2 with \textbf{FG}, although the difference is within the margin of error (0.6). Note that \newcite{kiela2018dynamic} report slightly different results than our reimplementation of their system, for \textbf{FG}: 86.2 $\pm$ .2 for SNLI (compared to 85.57 $\pm$ .3), and 88.7 $\pm$ .6 for SST2 (compared to 88.10 $\pm$ .6). These discrepancies are attributable to random initializations and the different representation of out-of-vocabulary words\footnote{While \newcite{kiela2018dynamic} uses zero-vectors to represent OOV words, we opted to use the facet-level centroid as it resulted in better validation performance for most tasks. We found we were unable to exactly replicate their results even with zero-vectors.}.

Our second finding is that concatenation is still a very strong baseline for meta-embeddings. This is not surprising because it preserves all of the information in the facets, and also introduces more model parameters, while the other meta-embedding methods seek to compress the information from all the meta-embeddings. Yet, for 4 out of the 6 tasks, word prisms outperform concatenation, and in the other two tasks (Semcor and SST2) word prisms obtain second best results within the margin of error. Furthermore, note that concatenation is very expensive in the case of including 13 facets, requiring 3900-dimensional meta-embeddings at inference time, versus only 300-dimensions for word prisms. 

Our third major finding is that sequence labelling tasks highly benefit from including all 13 facets, while the text classification tasks seem generally satisfied with only FG. Work on multi-task learning for supersense tagging on Semcor (using only a single set of embeddings with similar neural sequence labelling models) report results of 62.36 \cite{alonso2017multitask} and 68.25 \cite{changpinyo2018multi}. In contrast, our word prism model obtains a score of 73.82, indicating that word prisms can offer substantial improvements to supersense tagging models. Moreover, because supersense tagging is a coarse-grained version of word sense disambiguation, it is likely the case that word prisms can improve results in that domain as well. 
For further comparison, \newcite{huang2015bidirectional} obtain accuracies of 96.04 and 83.52 on the WSJ and NER tasks respectively (when using a BiLSTM with only a single set of embeddings as features) while our word prisms offer improvements to 97.04 (+1.00) and 90.74 (+7.22).   

\begin{figure}[t]
    \centering
    \includegraphics[trim=0 100 0 0, width=\linewidth]{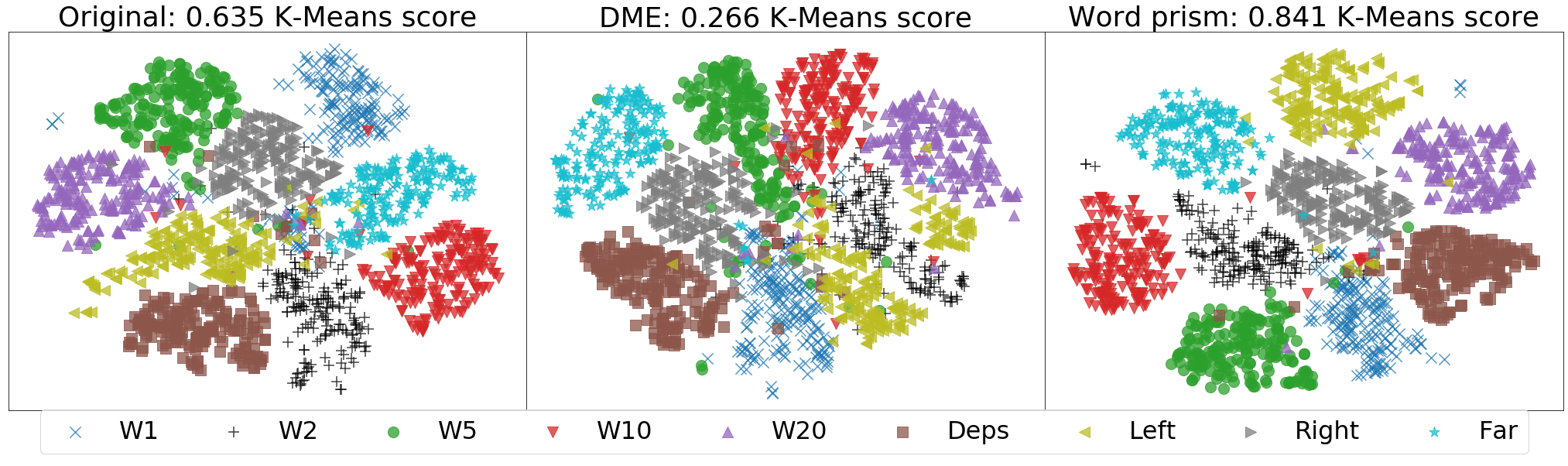}
    \caption{Impact of the linear transformations on the embedding space in dynamic meta-embeddings (DME) versus word prisms. The K-means score is the adjusted mutual information score for clustering embeddings into their respective facets, consistent across 5 clustering runs with different random seeds. Visualized is the TSNE-projected original facet-space versus the facets after projection (i.e., the stacked $\textbf{w}_f$ versus the stacked $\textbf{P}_f \textbf{w}_f + \textbf{b}_f$) in DMEs and word prisms.}
    \label{fig:projections}
\end{figure}

Our results additionally contribute to the findings of \newcite{coates2018frustratingly} on the difference between averaging and concatenation for meta-embeddings. When there are 13 facets, we observe a marked drop in quality for averaging. We posit that this is due to the increased noise pollution in the averaged vector space because, while it is likely that two sets of random vectors will be generally orthogonal, when 13 sets of random vectors come into play the ``birthday paradox'' becomes much more likely to unveil itself, and the odds of being well-separated become lower. 
In Figure~\ref{fig:projections}, we present the vector space for a subset of the nine window-based embeddings, projected to 2D with TSNE \cite{maaten2008visualizing}. Observe that the embeddings are not very well-separable on their own, as a K-Means clustering algorithm can only obtain an adjusted mutual information score of 0.635 to cluster them into their corresponding facets. We also present the embedding spaces being linearly transformed (i.e., $\textbf{P}_f \textbf{w}_f + \textbf{b}_f$, see Eq.~\ref{eq:meta}) when trained for supersense tagging. 
The learned, unconstrained transformations in DMEs cause the embedding space to become much less well-separated, as the clustering score deteriorates to 0.266. In contrast, the orthogonal transformations for word prisms improve the separatedness of the embedding space, bumping clustering score to 0.841. This (combined with the improved downstream results from word prisms) provides evidence of the validity of the ``natural clustering'' hypothesis for representation learning \cite{bengio2013representation,kenyon2019clustering}; namely, that it is preferable for neural representations to be well separated (or, disentangled) within their latent spaces.


\section{Ablation Studies}
\label{6_discussion}
We perform two ablation studies to further inspect the impact of the orthogonal transformations and meta-embedding combination choice in word prisms.  We compare the orthogonal transformation in word prisms to two alternatives: \textit{none} (i.e., $\textbf{P}_f = \textbf{I}_{d'}$, in which case the word prism is only learning the facet-combination weights $\alpha_f$), and an \textit{unconstrained} transformation which does not apply the orthogonality-imposing update rule detailed in Equation~\ref{eq:orthog}. For each experiment, we report the average and standard deviation across runs performed with five different random seeds.

\begin{table*}[t]
    \centering
        \begin{tabular}{c c c c c c c c}
            \hline
            Model & Proj. & Semcor & WSJ & Brown & NER & SNLI & SST2 \\
            \hline
            Avg. & None & 64.86 $\pm$ .2 & 96.53 $\pm$ .03 & 98.02 $\pm$ .02 & 86.33 $\pm$ .2 & 82.76 $\pm$ .2 & 85.55 $\pm$ .6\\
            Prism & None & 70.05 $\pm$ .2 & 96.73 $\pm$ .03 & 98.21 $\pm$ .02 & 87.56 $\pm$ .2 & 82.73 $\pm$ .4 & 85.61 $\pm$ .6 \\
            Prism & Uncon. & 71.54 $\pm$ .1 & 96.90 $\pm$ .01 & 98.36 $\pm$ .00 & 87.68 $\pm$ .3 & 83.87 $\pm$ .2 & 86.15 $\pm$ .7 \\
            Prism & Orthog. & \textbf{72.41} $\pm$ .2 & \textbf{96.98} $\pm$ .04 & \textbf{98.44} $\pm$ .01 & \textbf{88.91} $\pm$ .2 & \textbf{83.89} $\pm$ .4 & \textbf{86.68} $\pm$ .6 \\
            \hline
        \end{tabular}
    \caption{\textbf{Ablation study 1}: \textit{The transformation in word prisms}. Test set results for word prisms with different projection constraints (no transformation, an unconstrained one, and the orthogonal transformation), taking the nine window-based facets as the meta-embedding combination.}
    \label{tab:hilbert_projections}
\end{table*}
\begin{table*}[t]
    \centering
        \begin{tabular}{l c c c c c c}
            \hline
            Facets & Semcor & WSJ & Brown & NER & SNLI & SST2\\
            \hline
            Best-Window & 67.55 $\pm$ .1 
            & 96.60 $\pm$ .03 
            & 98.17 $\pm$ .01 
            & 87.54 $\pm$ .3 
            & 83.08 $\pm$ .2 
            & 85.82 $\pm$ .7 
            \\
            Best-All & 70.50 $\pm$ .2 
            & 96.75 $\pm$ .03 
            & 98.42 $\pm$ .03 
            & 90.39 $\pm$ .2 
            & 85.60 $\pm$ .2 
            & 87.48  $\pm$ 1.0  
            \\
            \hline
            W1-10 & 71.67 $\pm$ .2 & 96.86 $\pm$ .02 & 98.36 $\pm$ .02 & 88.39 $\pm$ .2 &   83.88 $\pm$ .3 & 86.52 $\pm$ .4\\
            W1-Far & 71.78 $\pm$ .1 & 96.89 $\pm$ .02 & 98.36 $\pm$ .01 & 88.54 $\pm$ .2 &   83.69 $\pm$ .3 & 87.39 $\pm$ .3 \\
            All windows & 72.41 $\pm$ .2 & 96.98 $\pm$ .02 & 98.44 $\pm$ .01 & 88.91 $\pm$ .2 &83.89 $\pm$ .4 & 86.68  $\pm$ .6 \\
            FG & 73.51 $\pm$ .1 & 96.91 $\pm$ .01 & 98.58 $\pm$ .02 & 89.70 $\pm$ .4 &   \textbf{85.82} $\pm$ .1 & 87.80 $\pm$ .6  \\
            FGCL & 73.51 $\pm$ .1 & 96.98 $\pm$ .03 & 98.63 $\pm$ .01 & 89.83 $\pm$ .3 &  85.68  $\pm$ .2 & \textbf{88.90} $\pm$ .4 \\
            All & \textbf{73.82} $\pm$ .2 & \textbf{97.04} $\pm$ .01 & \textbf{98.65} $\pm$ .02 & \textbf{90.74} $\pm$ .2 &   85.71 $\pm$ .3 & 88.45 $\pm$ .5  \\
            \hline
        \end{tabular}
    \caption{\textbf{Ablation study 2}: \textit{Different combinations of facets in word prisms.} Test set results when using different combinations of input facets in word prisms. First two rows contain only the single best performing window-based facet (Best-Window) or the best facet overall (Best-All), for the specific task.}
    \label{tab:facet_combos}
\end{table*}


\paragraph{Ablation study 1.} In this experiment we determine the impact of the transformation matrix in word prisms when isolating the input facets to be the 9 window-based facets trained on the same dataset, with the same vocabulary, differing only in the definition of the context window. 
Table~\ref{tab:hilbert_projections} presents the results for this experiment, which furthermore demonstrates the effectiveness of the orthogonal transformation.

\paragraph{Ablation study 2.} In this experiment we investigate the impact of different choices of facet combinations for word prisms. We experiment with several different sets of facets: \textbf{W1-10} denotes window sizes between 1 and 10 (inclusive) [4 facets]; \textbf{W1-Far} includes the prior facets plus \textit{W20} and \textit{Far} [6 facets]; \textbf{All windows} includes the \textit{Left}, \textit{Right}, and \textit{Deps} with the prior facets [9 facets]. 
\textbf{FG} denotes FastText and GloVe, while \textbf{FGCL} denotes FastText, GloVe, ConceptNet, and LexSub. 
\textbf{All} denotes all 13 of these facets. 
The results for these combinations are summarized in Table \ref{tab:facet_combos}.  In the first two rows, we also include the results for the best-performing (on the held out validation set) single-facet embedding model for the best window-based facet (\textbf{Best-Window}, out of the 9 window-based facets), and the best overall single-facet model (\textbf{Best-All} out of all 13 facets). \textit{W2} is the best window-based facet for Semcor, WSJ (tie), Brown, NER, and SST2, while \textit{W5} is the best window-based facet for WSJ (tie) and SNLI. GloVe is best overall for WSJ, Brown, NER, and SST2, while FastText is best overall for Semcor and SNLI.

We note three important takeaways from these results. First, word prisms always perform better than their single-facet counterparts, even though the single-facet models were selected to be the one that maximized validation performance for the specific task.  
Second, we observe that progressively incorporating more facets trained solely on different notions of context (i.e., from \textbf{W1-10} to \textbf{W1-Far} to \textbf{All windows}) improves results quality substantially for the sequence labelling tasks, while the text classification tasks (SNLI and SST2) do not benefit as much, although SST2 does seem to prefer the topic-based representations included by \textbf{W1-Far}. This suggests that NLP practitioners would benefit from training multiple sets of embeddings with different context windows in their specific problem domains (e.g., on a Twitter corpus), where they can expect improvements in results, especially if they are faced with a sequence labelling problem. 
Our third takeaway is that FastText and Glove are much better than the window-based embeddings, although including them all together still improves results in 3 out of the 6 tasks. This is not surprising since the FastText and Glove embeddings are trained on a corpus with over 600 billion words, while our window-based embeddings are only trained on a 6 billion word corpus; i.e., 1\% of the data of the former. Thus, our results in this ablation study and the former suggest that training embeddings with different notions of context on such corpora will lead to even further gains.

\section{Conclusion}
\label{7_conclusion}
In this paper, we study a simple and efficient method for constructing meta-embeddings from wide-ranging facets while preserving individual invariance with orthogonal transformations. 
The effectiveness of the proposed \textit{word prisms} is validated by six supervised extrinsic evaluation tasks. 
Our word prism models obtain consistent improvements over dynamic meta-embeddings \cite{kiela2018dynamic} and the averaging and concatenation baselines \cite{coates2018frustratingly} in all six tasks. 
Analysis of the transformed embeddings suggests the ``natural clustering'' hypothesis for representation learning \cite{bengio2013representation} is important to consider for combining various source embeddings to create performant task-specific meta-embeddings.


Several future directions present themselves from this work. First, we believe that contextualized embedding models can benefit from prismatic representations of their input embeddings \cite{devlin2019bert}, and that word prisms can benefit from including contextualized embeddings as facets.
Second, we expect that word prisms can improve performance in other tasks such as automatic summarization, which often use a single set of word embeddings in their input layers \cite{dong2019editnts}. Third, we believe that meta-embeddings and the method behind word prisms can be generalized past word-based representations to sentence representations \cite{pagliardini2018unsupervised} and may improve their quality, as was recently demonstrated by \newcite{poerner2019sentence}. 
Lastly, recent work has found simple word embeddings to be useful for solving diverse problems from the medical domain \cite{zhang2019biowordvec}, to materials science \cite{tshitoyan2019unsupervised}, to law \cite{chalkidis2019deep}; we expect that word prisms and their motivations can further improve results in these applications.

\section*{Acknowledgments}
This work is supported by the Fonds de recherche du Qu{\'e}bec – Nature et technologies, by the Natural Sciences and Engineering Research Council of Canada, and by Compute Canada. The last author is supported in part by the Canada CIFAR AI Chair program.


\bibliographystyle{coling}
\bibliography{coling2020}

\end{document}